\documentclass{article}

 \PassOptionsToPackage{numbers, compress}{natbib}
\usepackage[final]{nips_2017}

\usepackage[utf8]{inputenc} 
\usepackage[T1]{fontenc}    
\usepackage{hyperref}       
\usepackage{url}            
\usepackage{booktabs}       
\usepackage{amsfonts}       
\usepackage{nicefrac}       
\usepackage{subfiles}

\usepackage{natbib,float}
\usepackage{color}
\usepackage{algorithm}
\usepackage{algorithmic}

\usepackage{amsfonts,amsmath,amssymb,amsthm}
\usepackage{graphicx}
\DeclareGraphicsExtensions{.pdf,.png,.gif,.jpg}

\theoremstyle{plain}
\newtheorem{theorem}{Theorem}[section]

\theoremstyle{definition}

\theoremstyle{remark}

\providecommand{\delete}[1]{}
\providecommand{\nitro}[1]{}

\newcommand{\iid}{i.\@i.\@d.\@ }

\newcommand{\E}{{\mathbb E}}

\renewcommand{\P}{{\mathbb P}}

\newcommand{\cA}{{\cal A}}\newcommand{\cC}{{\cal C}}
\newcommand{\cD}{{\cal D}}\newcommand{\cF}{{\cal F}}
\newcommand{\cG}{{\cal G}}\newcommand{\cH}{{\cal H}}

\newcommand{\cM}{{\cal M}}
\newcommand{\cP}{{\cal P}}

\newcommand{\cX}{{\cal X}}\newcommand{\cZ}{{\cal Z}}

\newcommand{\N}{{\mathbb N}}\renewcommand{\P}{{\mathbb P}}
 \newcommand{\R}{{\mathbb R}}

\newcommand{\I}{{\mathbb I}}

\DeclareMathOperator*{\argmin}{arg\,min}

\newcounter{AssumptionCounter}

\title{Non-parametric estimation of \\Jensen-Shannon Divergence in \\Generative Adversarial Network training}

\author{
  Mathieu Sinn  \\
  IBM Research -- Ireland\\
  Mulhuddart, Dublin 15, Ireland\\
  \texttt{mathsinn@ie.ibm.com} \\
  \And
  Ambrish Rawat  \\
  IBM Research -- Ireland\\
  Mulhuddart, Dublin 15, Ireland\\
  \texttt{ambrish.rawat@ie.ibm.com} \\
}
\begin{document}

%

%

\maketitle

\begin{abstract}
Generative Adversarial Networks (GANs) have become a widely popular framework for generative modelling of high-dimensional datasets. However their training is well-known to be difficult.
This work presents a rigorous statistical analysis of GANs providing straight-forward explanations for common training pathologies such as vanishing gradients. Furthermore, it proposes a new training objective, Kernel GANs, and demonstrates its practical effectiveness on large-scale real-world data sets.  
A key element in the analysis is the distinction between training with respect to the (unknown) data distribution, and its empirical counterpart. 
To overcome issues in GAN training, we pursue the idea of smoothing the Jensen-Shannon Divergence (JSD) by incorporating noise in the input distributions of the discriminator.
As we show, this effectively leads to an empirical version of the JSD in which the true and the generator densities are replaced by kernel density estimates, which leads to Kernel GANs. 
\end{abstract}

\section{INTRODUCTION}
Generative Adversarial Networks (GANs), introduced by \citet{Goodfellow_GAN_original_paper_2014}, have become a widely popular framework for generative modeling using deep neural networks. 
While practitioners find that GANs -- particularly for image data -- produce sharp and realistic samples, it
is well recognized that GANs are difficult to train. Key challenges are: vanishing gradients, local optima leading to mode collapse, high sensitivity to hyperparameters, and finding the right balance between generator and discriminator training in the adversarial set-up (\citet{Dinh_Real_NVP_2016,Goodfellow_NIPS_tutorial_2016,Goodfellow_modified_GAN_2014,Metz_et_al_2016,Radford_2015,Goodfellow_modified_GAN_2016}).

Various authors have proposed practical modifications of GAN training to address these issues. However, only recently have authors begun to analyze them mathematically and develop principled solutions. An important step in this direction was the work by \citet{Arjovesky_Bottou_2017}, which led to the idea of Wasserstein GANs elaborated in \citet{Arjovesky_Chintala_Bottou_2017} and further developed by \citet{Gulrajani_et_al_2017}.
Two important insights were: 1) training the discriminator in GANs till optimality may provably result in vanishing gradients, and 2) the Jensen-Shannon Divergence (JSD) doesn't yield meaningful information about convergence of distributions if their intersection with the support of the limit-distribution has measure zero. Another important contribution was the work by \citet{Metz_et_al_2016}, who proposed to unroll discriminators in the GAN training objective in order to avoid degenerate optima and vanishing gradients.

\textbf{Our contributions.} This work has three major contributions. 

\begin{itemize}
	\item First, a rigorous mathematical framework to analyze GANs, which yields a remarkably simple explanation of the vanishing gradient problem.
	\item Second, a novel training objective, Kernel GANs, backed with a principled theoretical analysis along with an empirical study that highlights practical aspects of Kernel GAN training.
	\item Finally, experimentation with different training setups that scale Kernel GANs to large datasets and establish their practical usefulness. 
\end{itemize}

Our rigorous mathematical framework allows us to formulate the results in \citet{Arjovesky_Bottou_2017} more generally, and it addresses obscurities in the original theory developed in 
\cite{Goodfellow_GAN_original_paper_2014}. It appears that a major source of confusion has been the missing distinction between the GAN training objective with respect to the (unknown) target density, and its empirical counterpart used in practical training. This was recently independently pointed out by \citet{Arora_et_al_2017}, however, without deriving the implications stated in the present work.

To overcome GAN training pathologies, we analyze approaches for smoothing the JSD in the GAN training objective. 
We pursue the idea of adding noise terms to the inputs of the discriminator. As we show, this leads to an empirical version of the JSD in which the true and the generator densities are replaced by kernel density estimates. We establish almost sure convergence of this Kernel GAN objective and conditions under which it is asymptotically unbiased.

We support the theoretical analysis with extensive experimentation. Particular emphasis is on understanding the effect of the kernel bandwidth in the training algorithm. We also study the generalization of Kernel GANs by an evaluation of both the fidelity and the diversity of generated samples, also in comparison with baseline methods. Furthermore, we extend the Kernel GAN training setup to enable generative modelling of large-scale real-world datasets. 

\textbf{Related work.} In its attempt to establish a rigorous mathematical framework for understanding properties of GANs, this work is related to
\citet{Arjovesky_Bottou_2017}. However, the framework here is more general, e.g., many of the results cover both discrete and continuous distributions, and no parametric family assumptions are imposed on the generators. Moreover we clarify the consequences of working with empirical distributions in practical GAN training, which leads to a remarkably simple explanation of the pathologies discussed in \cite{Arjovesky_Bottou_2017}. 
Plugging kernel density estimates into the objective function bears some similarity with the work by \citet{Dziugaite:2015:TGN:3020847.3020875}, \citet{li2015generative} and, most recently, \cite{li2017mmdgan}. However, these authors optimize generative models with respect to the Maximum Mean Discrepancy (MMD) criterion. Contrastingly, our approach retains the adversarial setup; in fact, it can be regarded as unrolling the discriminator in the GAN training objective till optimality, which is similar in spirit to the methodology proposed by \citet{Metz_et_al_2016}.

\textbf{Outline of this paper.}
The mathematical framework and theoretical findings are established in Section \ref{sec:theoretical_analysis}, which also introduces the novel training objective. Section \ref{sec:Experiments} provides practical aspects of training with respect to that objective, and discusses experimental results. Section \ref{sec:Conclusions} concludes the paper. Proofs, implementation details and additional experiments are included in the Appendix.

\section{THEORETICAL ANALYSIS}\label{sec:theoretical_analysis}

\subsection{Preliminaries}\label{subsec:preliminaries}
Let $(\Omega, \cF, \P)$ be a probability space. Consider measurable spaces 
$(\cX,\cA)$ (the \textbf{output space}) and $(\cZ,\cC)$ (\textbf{latent space}). Let
$\cG$ denote the set of measurable functions $g: \cZ \to \cX$ (\textbf{generators}), and $\cD$ the set of
measurable functions $d:\cX \to [0,1]$ (\textbf{discriminators}). Let $\mu$ be a measure on $(\cX,\cA)$.
For the remainder of this paper, let $X$ and $Z$ be fixed random variable from $(\Omega, \cF, \P) $ onto $(\cX,\cA)$ and $(\cZ,\cC)$, respectively.
We will make frequent use of the following assumptions:
\begin{list}{(A\arabic{AssumptionCounter})}
{\usecounter{AssumptionCounter}}
\setcounter{AssumptionCounter}{0}
\item The distribution of $X$ is absolutely continuous with respect to $\mu$.
\item For every $g\in\cG$, the distribution of $g(Z)$ is absolutely continuous with respect to $\mu$.
\end{list}
As a consequence, $X$ has a $\mu$-density $p$
and $g(Z)$ has a $\mu$-density $p^{(g)}$
for every $g\in\cG$.
Given iid samples $X_1$, $X_2$, \ldots, $X_n$ from the distribution of $X$, our goal is to learn a generator $g$ such that $\P(g(Z) \in A) = \P(X \in A)$ for all $A\in\cA$, or, equivalently, $p=p^{(g)}$ $\mu$-almost everywhere. 
The next theorem establishes the existence of such a $g$ under the following assumptions:
\begin{list}{(A\arabic{AssumptionCounter})}
{\usecounter{AssumptionCounter}}
\setcounter{AssumptionCounter}{2}
\item $\cX$ is a Peano space, i.e., $\cX$ is a compact, connected, and locally connected metric space.
\item $\cX$ is the support of $X$, i.e., there doesn't exist an $x\in\cX$ with an open neighborhood $B_x$ in the topology of $\cX$ such that $\P(X\in B_x)=0$.
\end{list}

\begin{theorem}\label{theo:measure_preservation}
Suppose that {\rm(A1)-(A4)} hold. Moreover suppose that $\cZ=[0,1]$, $\cC$ is the Borel $\sigma$-algebra on $[0,1]$, and $Z$ is uniformly distributed on $\cZ$. Then there exists a continuous surjection 
$g: \cZ\to\cX$ such that $\P(g(Z) \in A) = \P(X \in A)$ for all $A\in\cA$.
\end{theorem}

Note that, equivalently, one could have assumed $Z$ follows any real-valued distribution which is absolutely continuous with respect to the Lebesgue measure.
There has been some confusion in the GAN literature about the exact conditions that are required to obtain this result. For example, 
\citet{Goodfellow_NIPS_tutorial_2016} states the ``the only requirements'' for $p^{(g)}$ to have ``full support'' on $\cX$ are that the dimension of $\cZ$ be ``at least as large as the dimension of $\cX$'', and $g$ be differentiable. This isn't accurate as $\cZ$ may have smaller dimension, as long as its cardinality is not smaller than the one of $\cX$, and the distribution of $Z$ is non-atomic.
Differentiability of $g$ is not required in theory. To obtain an invertible and differentiable mapping $g$, the dimension of $\cZ$ must not be smaller than the dimension of $\cX$. The result in Theorem \ref{theo:measure_preservation} relies on a construction using space-filling curves, which aren't differentiable.

\subsection{GAN Training -- Theoretical Case}\label{subsec:population_case}

The GAN approach (\citet{Goodfellow_GAN_original_paper_2014}) for learning $g$ is as follows: for $d\in\cD$ and $g\in\cG$ let
\begin{eqnarray}\label{eq:GAN_objective}
V(d,g) \,:=\, \E\big[\log(d(X))\big] + \E\big[\log(1-d(g(Z))) \big].
\end{eqnarray}
The relation of $V(d,g)$ to density ratio estimation (which becomes apparent in equation (\ref{eq: prop1.2}) below) is discussed in \cite{Mohamed_Lakshminarayanan_2016}. 
Intuitively, we wish the discriminator $d(x)$ to be close to $1$ if $x$ is more likely under the distribution of $X$, and close to $0$ if $x$ is more likely under the distribution of $g(Z)$. Hence, the optimal $d$ given a fixed generator $g$ would attempt to maximize $V(\cdot,g)$, and the optimal $g$ is the one which solves the minmax problem
\begin{eqnarray}\label{eq:def_optimal_generator}
g &=& \argmin_{g\in\cG} \Big( \max_{d\in\cD} V(d,g) \Big).
\end{eqnarray}
The following theorem, which generalizes Proposition 1 and Theorem 1 in \cite{Goodfellow_GAN_original_paper_2014}, shows that the max and (arg)min in (\ref{eq:def_optimal_generator}) are well-defined. Note that our formulation neither requires $g$ to be differentiable, nor $\cX$ to be continuous.

\begin{theorem}\label{theo:population} Suppose {\rm(A1)-(A2)} hold. Then
\begin{eqnarray}\label{eq: prop1.1}
V(d,g) &=&  \int_{\cX} \Big[ \log(d(x)) p(x)  \nonumber\\ && \hspace{3mm} \,+\, \log(1-d(x)) p^{(g)}(x)  \Big] \mbox{\rm d}\mu(x)
\end{eqnarray}
 for all $d\in\cD$ and $g\in\cG$. Hence, for any fixed $g\in\cG$, any $d\in\cD$ which maximizes $V(g,d)$ has the form
\begin{eqnarray}\label{eq: prop1.2}
d(x) &=& \frac{p(x)}{p(x) + p^{(g)}(x)}
\end{eqnarray}
for $\mu$-almost every $x\in\cX$, implying that
\begin{eqnarray}\label{eq: prop1.3}
\max_{d\in\cD} V(g,d) &=&  \int_{\cX} \Big[ p(x) \log  \frac{p(x)}{p(x) + p^{(g)}(x)}  \nonumber\\ && \hspace{-10mm}  \,+\, p_{g}(x) \log \frac{p^{(g)}(x)}{p(x) + p^{(g)}(x)} \Big] \mbox{\rm d}\mu(x).
\end{eqnarray}
Assuming that {\rm(A3)-(A4)} also hold, any generator $g\in\cG$ that minimizes {\rm (\ref{eq: prop1.3})} is such that $p^{(g)}=p$ $\mu$-almost everywhere, and $\min_{g\in\cG} \max_{d\in\cD} V(g,d) = -\log(4)$. 
\end{theorem}

The next theorem establishes further properties of the optimal discriminator $d$ in (\ref{eq: prop1.2}). It generalizes Theorem 2.1 and 2.2 in \cite{Arjovesky_Bottou_2017}, which were stated for the special case of $\cP$ and $\cP^{(g)}$ being not-perfectly-aligned submanifolds of $\R^k$.

\begin{theorem}\label{theo:optimal_discriminator}
Suppose {\rm(A1)-(A3)} hold. 
For fixed $g\in\cG$, let $\cP,\cP^{(g)}\subset\cX$ be such that 
$\{x\in\cX\,|\,p(x)>0\}\subset\cP$ and $\{x\in\cX\,|\,p^{(g)}(x)>0\}\subset\cP^{(g)}$. 
Suppose $\mu(\cP \cap \cP^{(g)})=0$, $\mu(\partial(\cP \setminus \cP^{(g)}))=0$ (where $\partial(\cdot)$ denotes the topological boundary)
and $\mu(\partial( \cP^{(g)} \setminus \cP))=0$. Then the optimal
$d$ in {\rm (\ref{eq: prop1.2})} satisfies $\P(d(X)=1)=1$ and $\P(d(g(Z))=0)=1$. 
Moreover, without loss of generality, $d$ is continuous $\mu$-almost everywhere and, in the special case $\cX=\R^k$, 
the gradient $\nabla d(x)$ exists and $\nabla d(x) = 0$ for $\mu$-almost every $x\in\cX$.
\end{theorem}

In practice, the discriminator being constant on $\cP$ and $\cP^{(g)}$ poses problems.
In particular, when $g$ is fixed and $d$ is trained till optimality, the gradients $\nabla d(x)$ may vanish and further updates of $g$ become impossible.
In their Lemma 1 and 2, \citet{Arjovesky_Bottou_2017} establish that this is almost surely going to occur whenever the 
dimension of $\cZ$ is smaller than the dimension of $\cX$, and $g$ is parameterized by a standard neural network. 
As we show next, it is more directly an inevitable consequence of using an empirical version of the objective (\ref{eq:GAN_objective}) in practical GAN training.

\subsection{GAN Training -- Empirical Case}\label{subsec:empirical_case}

Let $X_n^\ast$ be a random variable following the empirical distribution of $X_1$, \ldots, $X_n$. By $\I(\cdot)$ we denote the indicator function which evaluates to $1$ if the statement in brackets is true, and to $0$ otherwise. Note that, conditionally on $X_1$, \ldots, $X_n$, the distribution of  $X_n^\ast$ is 
\begin{eqnarray*}
\P(X_n^\ast \in A \, | \, X_1,\ldots,X_n) &=& \frac{1}{n}\sum_{i=1}^n \I(X_i \in A)
\end{eqnarray*}
for $A\in\cA$, and an analogous statement holds for the distribution of $g(Z_n^\ast)$ conditional on $Z_1$, \ldots, $Z_n$. It is important to note that practical GAN training (such as in Algorithm 1 in \cite{Goodfellow_GAN_original_paper_2014}) is {\it not} with respect to the theoretical objective (\ref{eq:GAN_objective}), but with respect to its empirical counterpart
\begin{eqnarray}\label{eq:empirical_objective}
&&  \hspace{-5mm} V_n(d,g) \,:=\, \E\big[
\log(d( X_n^\ast ) ) \, | \, X_1,\ldots,X_n
\big]  \nonumber\\ && \hspace{10mm} \, + \, 
\E\big[
\log(1 - d(g(Z_n^\ast) ) ) \, | \, Z_1,\ldots,Z_n
\big] .
\end{eqnarray}
It appears there has been a wideheld belief among GAN practitioners that optimizing $V_n(d,g)$ leads to discriminators and generators with the same properties as stated in Theorem \ref{theo:optimal_discriminator}. As the following theorem shows, this isn't true in general. 
We add subscripts $d_n$ and $g_n$ to emphasize the dependency of discriminator and generator on the sample size $n$.

\begin{theorem}\label{theo:empirical_case}
Suppose {\rm(A1)-(A4)} hold. 
For fixed $g\in\cG$, any $d_n\in\cD$ maximizing $V_n(d,g)$ in {\rm (\ref{eq:empirical_objective})} has the form
\begin{eqnarray}\label{eq:optimal_empirical_discriminator}
d_n(x) \,=\, \frac{ \sum_{i=1}^n \I(X_i=x) }{ \sum_{i=1}^n \I(X_i=x)  +  \sum_{i=1}^n \I(g(Z_i)=x) }
\end{eqnarray}
for $x\in\{X_1,\ldots,X_n\}\cup\{g(Z_1),\ldots,g(Z_n) \}$ (for all other $x\in\cX$, the value $d_n(x)$ is arbitrary).
If the cardinality of $\{Z_1$, \ldots, $Z_n$\} is greater than or equal to the cardinality of $\{X_1$, \ldots, $X_n\}$, then any $g_n\in\cG$ minimizing {\rm (\ref{eq:empirical_objective})} for $d=d_n$ is such that
$\{g_n(Z_1),\ldots,g_n(Z_n) \}=\{X_1,\ldots,X_n\}$.
\end{theorem}

Theorem \ref{theo:empirical_case} reveals two insights: First, if $X$ and $Z$ both have continuous distributions, then $d_n$ has the same properties as $d$ in Theorem \ref{theo:optimal_discriminator}. This suggests the primary reason for vanishing gradients in  GAN training is the discrete nature of the empirical objective (\ref{eq:empirical_objective}) -- which is a remarkably simple explanation. 

The second insight is that, when training with respect to (\ref{eq:empirical_objective}), there is no theoretical guarantee that $p^{(g_n)}=p$ $\mu$-almost everywhere for the optimal generator $g_n$ -- which contradicts Proposition 2 in \cite{Goodfellow_GAN_original_paper_2014}. The only guarantee is that, when applied to $Z_1, \ldots, Z_n$, $g_n$ should reproduce the training samples $X_1,\ldots,X_n$.
Note: this does not imply that $g_n$ will {\it solely} reproduce training samples; in theory, the samples generated on $\cZ\setminus\{Z_1,\ldots,Z_n\}$ are arbitrary. Hence, in contrary to the reasoning in \cite{Metz_et_al_2016} and \cite{Arjovesky_Chintala_Bottou_2017}, the optimal $g_n$ is {\it not} necessarily a Dirac function at the $x\in\cX$ to which $d_n$ assigns the highest values.

In practice, these undesirable properties could be mitigated for the following reasons: 
1) the discriminator and generator function spaces $\cD$ and $\cG$ have limited capacity, hence the properties of $d_n$ and $g_n$ may only hold approximately; 2) similarly, alternate training of the generator and discriminator, or not training till optimality could alter the form of $d_n$ and $g_n$, thereby circumventing pathologies.
Limiting the capacity of the networks or finding the right balance between training the generator and discriminator, however, is challenging. This is why GAN training has been regarded as extremely difficult among practitioners.

\subsection{Smoothing the Training Objective}\label{subsec:smoothing}

A natural approach to avoid the issues pointed out in Theorem \ref{theo:optimal_discriminator} and Theorem \ref{theo:empirical_case} is to smooth the Jensen-Shannon Divergence (JSD) in the GAN training objective by adding noise to the input distributions of the optimal discriminator.\footnote{This was previously discussed in \cite{Arjovesky_Bottou_2017}, however, the idea was not pursed beyond an initial analysis.} In the following, let $\epsilon$ be a fixed random variable on $(\cX,\cA)$ which is absolutely continuous with respect to $\mu$, hence $\epsilon$ has a $\mu$-density $p^{(\epsilon)}$.
We use the following assumption:
\begin{list}{(A\arabic{AssumptionCounter})}
{\usecounter{AssumptionCounter}}
\setcounter{AssumptionCounter}{4}
\item In addition to (A3), $(\cX,+)$ is a topological group.
\end{list}
This allows us to consider the convolutions $p\ast p^{(\epsilon)}$ and $p^{(g)}\ast p^{(\epsilon)}$, which are the $\mu$-densities of
$X+\epsilon$ and $g(Z)+\epsilon$, respectively.
The idea is to use, instead of the discriminator in (\ref{eq: prop1.2}), a modified version
\begin{eqnarray}\label{eq:modified_optimal_discriminator}
d^\ast(x) &=&  \frac{p\ast p^{(\epsilon)}(x)}{p\ast p^{(\epsilon)}(x) + p^{(g)}\ast p^{(\epsilon)}(x)}.
\end{eqnarray}
If the support of $p^{(\epsilon)}$ is sufficiently large, then the supports of $p\ast p^{(\epsilon)}$ and
$p^{(g)}\ast p^{(\epsilon)}$ will overlap. Hence, it is not possible to construct an optimal $d^\ast$ with
the properties in Theorem \ref{theo:optimal_discriminator}.
On the other hand, by the same arguments as in Theorem \ref{theo:population}, the generator $g$ minimizing
(\ref{eq:modified_optimal_discriminator}) is such that  $p^{(g)}\ast p^{(\epsilon)} = p\ast p^{(\epsilon)}$ $\mu$-almost everywhere, which implies $p^{(g)}=p$ $\mu$-almost everywhere, i.e.~the optimal generator $g(Z)$ with respect to the theoretical objective still recovers the distribution of $X$. 
Next, we derive the form of the optimal discriminator for the modified empirical objective.

\begin{theorem}\label{theo:optimal_modified_empirical_discriminator} Suppose {\rm (A1)-(A5)} hold and let
$g\in\cG$ be fixed.
If we replace $X_n^\ast$ and $g(Z_n^\ast)$ in {\rm (\ref{eq:empirical_objective})} by
$X_n^\ast+\epsilon$ and $g(Z_n^\ast)+\epsilon$, respectively, then the discriminator minimizing the objective has the form
\begin{eqnarray}\label{eq:optimal_modified_empirical_discriminator}
d_n^\ast(x) &=&  \nonumber\\ && \hspace{-15mm} \frac{ \sum_{i=1}^n p^{(\epsilon)}(x-X_i)}
{  \sum_{i=1}^n p^{(\epsilon)}(x-X_i) \, + \,  \sum_{i=1}^n p^{(\epsilon)}(x-g(Z_i)) }
\end{eqnarray}
for $x\in\cX$. Same as in Theorem \ref{theo:empirical_case}, if the cardinality of $\{Z_1, \ldots, Z_n\}$ is greater than or equal to the cardinality of $\{X_1, \ldots, X_n\}$, then any $g_n^\ast\in\cG$ minimizing the objective {\rm (\ref{eq:empirical_objective})} for $d=d_n^\ast$ is such that
$\{g_n^\ast(Z_1),\ldots,g_n^\ast(Z_n) \}=\{X_1,\ldots,X_n\}$.
\end{theorem}

Note that the smoothing of distributions outlined here is {\it not} equivalent to adding noise to the samples 
$X_1$, \ldots, $X_n$ or $g(Z_1)$, \ldots, $g(Z_n)$ before optimizing the empirical objective,
which would lead to the same result as in (\ref{eq:optimal_empirical_discriminator}).

As Theorem \ref{theo:optimal_modified_empirical_discriminator} shows, smoothing the empirical distributions $X_n^\ast$ and $g(Z_n^\ast)$ results in
an optimal discriminator $d_n^\ast$ which, if the support of $p^{(\epsilon)}$ is sufficiently large, won't cause vanishing gradients. However, there is still no guarantee that the optimal generator $g_n^\ast(Z)$ recovers the distribution of $X$ apart from reproducing training samples.
In the following section we discuss a new training objective which addresses this issue.

\subsection{Kernel GANs}\label{sec:Kernel_GANs}

Throughout the rest of the paper we assume $\cX = \R^k$, $\mu$ is absolutely continuous with respect to the Lebesgue measure on $\R^k$, $\cZ\subset\R^l$, and $\Theta \subset \R^m$ for some $k,l,m\in\N$. Moreover, we assume that $\cG$ is parameterized by $\theta\in\Theta$. We write $g_\theta$ for the generator parameterized by $\theta$, and $p^{(\theta)}$ for the density of $g_\theta(Z)$.

It is instructive to note the resemblance of the optimal discriminator 
$d_n^\ast$ in (\ref{eq:optimal_modified_empirical_discriminator}) with a ratio of kernel densities:
Let $K:\cX\to\R$ be a measurable, bounded and square-integrable function ({\bf kernel}), and $\sigma>0$ ({\bf bandwidth}).
For $x\in\cX$ consider the kernel density estimates
\begin{eqnarray}\label{def:kernel1}
\hat{p}_{n,\sigma}(x) &:=& \frac{1}{\sigma^k n} \sum_{i=1}^n  K\left(\frac{x-X_i}{\sigma}\right), \\\label{def:kernel2}
\hat{p}_{n,\sigma}^{(\theta)}(x) &:=& \frac{1}{\sigma^k n} \sum_{i=1}^n   K\left(\frac{x-g_\theta(Z_i)}{\sigma}\right) 
\end{eqnarray}
of $p(x)$ and $p^{(\theta)}(x)$. Choosing $K=p^{(\epsilon)}$, we can regard (\ref{eq:optimal_modified_empirical_discriminator}) as a kernel estimate of the density ratio $p(x)/(p(x)+p^{(\theta)}(x))$.
Our key idea is to plug the optimal discriminator (\ref{eq:optimal_modified_empirical_discriminator}) back into the empirical training objective 
(\ref{eq:empirical_objective}), i.e., consider $V_n(d,g)$ with $d=d_n^\ast$. This results in the {\bf Kernel GAN} training objective: 
\begin{eqnarray}\label{eq:Kernel_GANs_parameter}
&& \hspace{-15mm} K_n(\theta, \sigma, \varphi) \,:=\,  \nonumber\\
&& \hspace{-09mm}
\frac{1}{n} \sum_{i=1}^n \log \frac{\hat{p}_{n,\sigma}(X_i)\,+\, \varphi}{\hat{p}_{n,\sigma}(X_i) \,+\, \hat{p}_{n,\sigma}^{(\theta)}(X_i)\,+\,2 \varphi} \nonumber\\ && \hspace{-13mm} \, +\,
\frac{1}{n} \sum_{i=1}^n \log \frac{\hat{p}_{n,\sigma}^{(\theta)}(g_\theta(Z_i))\,+\, \varphi}{\hat{p}_{n,\sigma}(g_\theta(Z_i))\,+\, \hat{p}_{n,\sigma}^{(\theta)}(g_\theta(Z_i))\,+\,2 \varphi} 
\end{eqnarray}
where $\varphi\geq 0$ is a regularizer to avoid underflow issues. 
In contrast to conventional GAN training, only the generator is explicitly updated when optimizing (\ref{eq:Kernel_GANs_parameter}).
The discriminator $d_n^\ast$ is updated implicitly through changes in the density estimates (\ref{def:kernel2}).
Note that plugging the optimal discriminator in (\ref{eq:optimal_modified_empirical_discriminator}) into the training objective (\ref{eq:empirical_objective}) can be regarded as unrolling the discriminator same as in 
\citet{Metz_et_al_2016}, where in our case the discriminator is unrolled to closed-form optimality.

 
The following theorem establishes convergence of the objective (\ref{eq:Kernel_GANs_parameter}).

\begin{theorem}\label{theo:consistency}
Suppose {\rm (A1)-(A5)} hold. Moreover, suppose $p$ and $p^{(\theta)}$ are bounded and uniformly continuous for all $\theta\in\Theta$, and $K$ has compact support and is of the form $K(x)=\phi(q(x))$, where $q$ is a polynomial and $\phi$ a bounded non-negative function with bounded variation. Let $\sigma_n>0$ be a sequence asymptotically equivalent to $Cn^{\frac{\delta-1}{k}}$ for some finite constant $C$ and $\delta\in(0,1)$. Then
\begin{eqnarray}\label{eq:theo_consistency}
&& \hspace{-18mm} \lim_{n\to\infty} K_n(\theta, \sigma_n, \varphi) \,=\,  \nonumber\\
&& \hspace{-11mm} \int_{\cX} \Big[ p(x) \log  \frac{p(x)\,+\,\varphi}{p(x) \,+\, p^{(\theta)}(x)\,+\,2 \varphi}  \nonumber\\
&& \hspace{-08mm}  \,+\, p^{(\theta)}(x) \log \frac{p^{(\theta)}(x)\,+\,\varphi}{p(x)\,+\, p^{(\theta)}(x)\,+\,2 \varphi} \Big] \mbox{\rm d}\mu(x)
\end{eqnarray}
$\P$-almost surely for all $\theta$ and $\varphi>0$. 
\end{theorem}

The regularizer $\varphi>0$ is required for establishing the convergence in (\ref{eq:theo_consistency}).
It results in estimates of the theoretical JSD that are asymptotically biased. 
In particular, while $K_n(\theta,\sigma_n,\varphi)$ converges to $-\log(4)$ (which is the minimum value of JSD) if $\theta$ is such that $p^{(\theta)}=p$, it may converge to smaller values for other values of $\theta$. Hence, minimizing $K_n(\theta,\sigma_n,\varphi)$ would not result in a generator $g^{(\theta)}(Z)$ recovering $X$ (although $\varphi$ can be chosen arbitrarily small, hence the practical difference might be negligible). 
However, as we show in Appendix A.2, if $\mu(\cX)<\infty$, then $K_n(\theta, \sigma_n, \varphi)$ can be modified such that its limit is minimized by a $\theta$ recovering the distribution of $X$.



\section{EXPERIMENTS}\label{sec:Experiments}

In this section, we discuss practical learning of Kernel GANs. First, we demonstrate the learning on small and mid-sized datasets -- a Mixture-of-Gaussian (MOG) toy dataset (\cite{Metz_et_al_2016}) and MNIST (\cite{lecun1998gradient}). Further, we study the effect of kernel bandwidth along with practical approaches such as generating in a lower-dimensional feature space that is independently learned using an autoencoder. 

Second, we establish practical usefulness of Kernel GANs by scaling them to two high-dimensional datasets: CIFAR-10 (\cite{Krizhevsky2009CIFAR}) and CelebA (\cite{Liu2015CelebA}). We enable this with a modified training setup that involves kernel learning, similar to \cite{li2017mmdgan}. 

Finally, we conduct various evalutions of the performance of the trained generators. In a quantitative evaluation, we compare Kernel GANs with MMD-based models (\citet{li2015generative}, \citet{Dziugaite:2015:TGN:3020847.3020875}, \citet{li2017mmdgan}), which also use kernel-based statistics, but in a non-adversarial fashion. Full details on the implementation and all the experiments can be found in Appendix A.3 and A.4.

\subsection{Learning Kernel GANs} 
\label{sub:learning_kgan}


Algorithm \ref{algo:KGAN} outlines our general training protocol for learning the generator parameters $\theta$ that minimize the training objective (\ref{eq:Kernel_GANs_parameter}). 

\begin{algorithm}
   \caption{Training Protocol}
   \label{algo:KGAN}
\begin{algorithmic}[1]
\STATE {\bf Input:}
Training samples $\boldsymbol{X}$, distribution of latent variable $Z$, initial kernel parameters $\sigma$, initial generator parameters $\theta$, regularizer $\varphi$.
\WHILE{{\it stopping criterion not met}}
\STATE Sample a minibatch $X_1$, \ldots, $X_n$ from $\boldsymbol{X}$.
\STATE Generate iid samples $Z_1$, \ldots, $Z_n$ from $Z$.
\STATE Update generator parameters $\theta$ according to gradients $\nabla_\theta K_n(\theta,\sigma,\varphi)$.
\STATE Update kernel parameters $\sigma$.
\ENDWHILE
\STATE {\bf Output:} Trained generator parameters $\theta$.
\end{algorithmic}
\end{algorithm}

\paragraph{Hyperparameters.}
Previously, \citet{li2015generative} and \citet{Dziugaite:2015:TGN:3020847.3020875} used RBF kernels in their training objectives for generative models. While \cite{li2015generative} deploys a mixture of RBF kernels, \cite{Dziugaite:2015:TGN:3020847.3020875} uses Bayesian Optimisation to determine a suitable bandwidth. Moreover, both works suggest to use the median-trick (\citet{gretton2012kernel}) as a method to choose kernel bandwidths for computing MMD statistics. 
Intuitively, small bandwidths push the generator towards producing samples that are similar 
to the training set.
However, initial bandwidths that are too small will not give gradients in areas that are far from the modes of the training set.
We therefore explored gradual reductions of the bandwidth during training, similar to cooling schedules in simulated annealing (e.g., \citet{Hajek_1988}, \citet{Nourani_Andresen_1998}).
While the regularizer $\varphi$ is required for deriving Theorem \ref{theo:consistency}, we didn't find it to play a crucial role in the practical experiments and therefore set it equal to zero. A further investigation of the practical effect of $\varphi$ will be part of future work.

\paragraph{MOG Toy Dataset.}
For the MOG dataset, $Z$ was a 100-dimensional standard normal distribution, and the generator used was a three-layer fully connected network (128-relu-128-relu-128-tanh).
Figure 1 shows the evolution of the generator during the training, as the bandwidth $\sigma$ is gradually decreased. Initially the generated samples $g(Z)$ are dispersed randomly. As the bandwidth $\sigma$ is decreased, they begin to concentrate around the modes of the MOG distribution. 

\begin{figure*}
\centering
\label{fig:evolution}
\includegraphics[width=\textwidth]{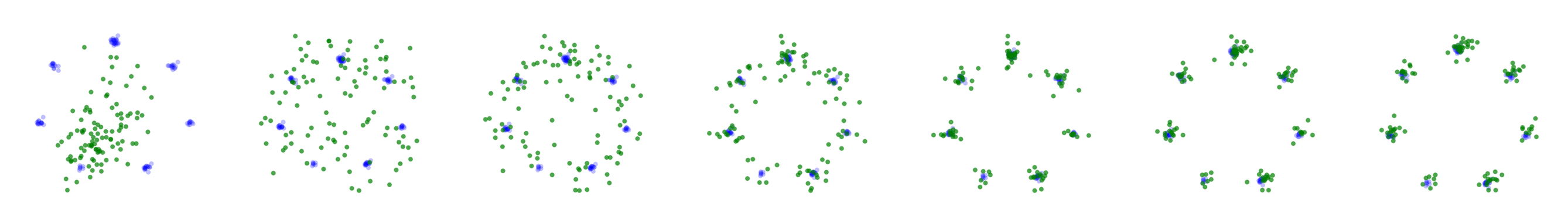}
\caption{MOG toy dataset. Blue: training points. Green: samples produced by the generator. The \textit{leftmost} figure is for the initial generator. Training phases for bandwidths (\textit{left} to \textit{right}): $0.8, 0.4, 0.2, 0.1, 0.05, 0.025$ (10,000 iterations were performed for each $\sigma$).}
\end{figure*}

\paragraph{MNIST.}
We succesfully trained three different generative models for MNIST. Two of these were trained to sample directly in the space of $(28\times28)$ greyscale images. The third model used an autoencoder to map the images onto a lower-dimensional feature space, in which the generator was trained.

The three models used following architectures: a fully connected network ({\bf FC}); a deconvolutional network with batch normalisation ({\bf DC}); a fully connected network for the feature space ({\bf FC-FS}). We adopted the architectures proposed in \cite{li2015generative} for FC and FC-FS, and the architecture proposed in \cite{Radford_2015} for DC. As latent variable $Z$, all models used samples from a 10-dimensional uniform distribution. We used a mixture of RBF kernels for training these models (see the appendix for details). For FC-FS, we also experimented with different bandwidths in simple RBF kernels. 

\begin{figure*}[t]
\begin{picture}(300,160)
\put(75,150){\small (a)}
\put(240,150){\small (b)}
\put(405,150){\small (c)}
\put(5,0){\includegraphics[width=5cm]{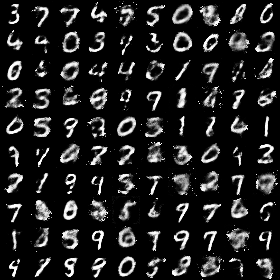}}
\put(170,0){\includegraphics[width=5cm]{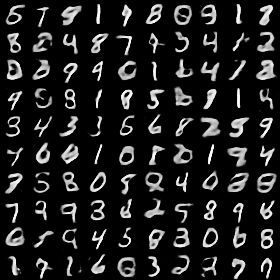}}
\put(335,0){\includegraphics[width=5cm]{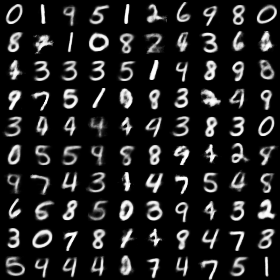}}
\end{picture}
\caption{Training of generators for MNIST. (a): Fully connected network (FC). (b): Deep convolutional architecture (DC). (c): Fully connected network in feature space (FC-FS).}
\label{fig: MNIST}
\end{figure*}
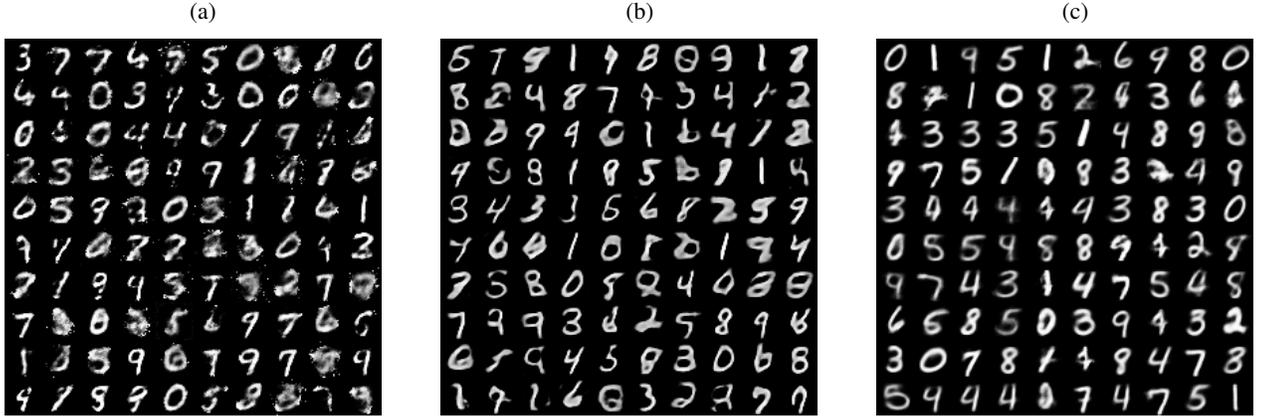

Generated samples from FC, DC and FC-FS are shown in Figure \ref{fig: MNIST} (a)-(c). 
We found that samples from FC-FS had a very smooth appearance. DC generated sharper samples than FC, but still produced some artifacts. The sharpness of the FC-FS samples with simple RBF kernels and different bandwidths is evaluated in the appendix. Quantitative measures of sample fidelity and diversity are discussed below.

We observed that training randomly initialized networks can be numerically unstable for very small bandwidths, leading to artifacts in the produced images. For very large bandwidths, we occasionally found the generator to collapse and produce undesired samples like mean images. 
We noticed, however, that the generator model recovered when we increased or decreased the bandwidth appropriately in subsequent training iterations. An analysis is provided in the appendix. This suggests that kernel bandwidths can be used as ``knobs'' for correcting over- or underfitting of generative models during the training process.



\begin{figure*}[t]
\begin{picture}(300,160)
\put(130,150){\small (a)}
\put(350,150){\small (b)}
\put(60,0){\includegraphics[width=5cm]{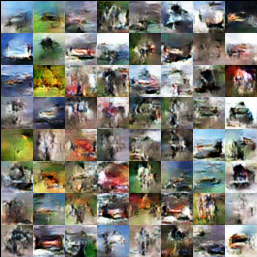}}
\put(280,0){\includegraphics[width=5cm]{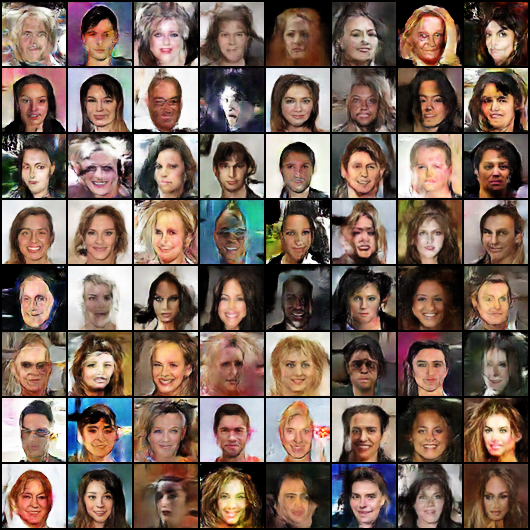}}
\end{picture}
\caption{Kernel GAN generated samples. (a): CIFAR10. (b): CelebA.}
\label{fig: colored}
\end{figure*}

\begin{table*}
  \caption{Quantitative evaluation of different generators on MNIST}
  \label{tab:class_mean2}
  \centering
\begin{tabular}{lrrrrrr}
\toprule
                           &   FC &   DC & FC-FS &   GMMN-AE &  GMMN & MNIST Test\\
\midrule
\textbf{EE}     &  0.408  &  0.289 &  0.365  &  0.361  & 0.293 & 0.023  \\
\textbf{ENN}      &  23.005  & 20.713 & 14.674  &  14.730  & 27.852 &19.299 \\
\textbf{LS}              &   6.601    &  7.464 & 6.916   &  6.948 & 7.408 &  9.752 \\
\textbf{JSD}              &   -1.384    & -1.385  & -1.371   &  -1.372 & -1.383 &  -1.386\\
\textbf{JSD-F}               &   -0.740    & -0.740 & -0.769   &  -0.776 & -0.689 &  -0.693 \\
\textbf{JSD-S}               &   -0.644    & -0.645 &  -0.603 &  -0.596  & -0.694 &-0.693 \\
\textbf{MMD}              &   0.019    &  0.017 & 0.084 &  0.091  & 0.005 &0.000 \\
\bottomrule
\end{tabular}
\end{table*}

\subsection{Scaling Kernel GANs} 
\label{sub:scaling_kernel_gans}

Although being sufficient in theory, we found it difficult to train Kernel GANs for colored images using plain RBF kernels. To impose more structure in kernel-based training of generative models, \cite{li2015generative} had suggested to use convolutional autoencoders to learn a lower dimensional feature space for colored images. \cite{li2017mmdgan} used this approach for training GMMN models on colored images both in feature and data space, however they observed quality issues in the generated samples. Instead they pursued an alternative approach where, much like a GAN setup, they learn a network which transforms the original space into a lower dimensional space over which the kernel is computed. We adopt their approach, leading to a modified Kernel GAN training objective, $K_n(\psi,\theta, \sigma, \varphi)$, which is the same as  (\ref{eq:Kernel_GANs_parameter}), except that the kernels operate on the space $f_\psi(\cX)$ instead of $\cX$:
\begin{eqnarray}\label{def:kernel1mod}
\hat{p}_{n,\sigma}^{(\psi)}(x) \,:=\, \frac{1}{n} \sum_{i=1}^n  K\left(\frac{f_\psi(x)-f_\psi(X_i)}{\sigma}\right), &&\\\label{def:kernel2mod}
\hat{p}_{n,\sigma}^{(\theta,\psi)}(x) \,:=\, \frac{1}{n} \sum_{i=1}^n   K\left(\frac{f_\psi(x)-f_\psi(g_\theta(Z_i))}{\sigma}\right). &&
\end{eqnarray}
The parameters $\{\theta,\psi\}$ are learned in a min-max fashion: $\text{min}_\theta\text{max}_\psi K_n(\psi,\theta, \sigma, \varphi)$. Similar to conventional GAN training, $\psi$ and $\theta$ are optimized alternatingly. In practice, we observed that additional regularization was required for stable learning. We used the experimental setup of \cite{li2017mmdgan}, which models the function $f_\psi$ as the encoder of an autoencoder and regularizes the objective function with the autoencoder reconstruction loss.

\paragraph{CIFAR10 and CelebA.}
We used this setup to successfully train Kernel GANs for the CIFAR10 and CelebA datasets. We adopted and appropriately rescaled hyperparameters and regularization weights of \cite{li2017mmdgan}.
We trained a Deep Convolutional architecture for both datasets. The dimension of the encoded space $f\psi(\cX)$ was fixed to 100. While CIFAR10 was trained with a 128-dimensional standard normal distribution for $Z$, CelebA was trained with a 64-dimensional $Z$. Samples obtained from the trained generators are shown in Figure \ref{fig: colored}. We found that they were qualitatively comparable to the results in \cite{li2017mmdgan}.

\subsection{Quantitative Evaluation}\label{sub:evaluation}

\paragraph{MNIST.}
Quantifying the performance of generative networks -- particularly their ability to generalize and produce diverse samples --
remains a challenging task (\cite{Theis_et_al_GAN_evaluation_2015, Wu_et_al_2017}).
In this paper, we report the following metrics: 

Expected entropy ({\bf EE}): As proposed in \cite{Goodfellow_modified_GAN_2016}, we trained a probabilistic 
classifier (LeNet \cite{lecun2015lenet}) and computed the
expected entropy of the classifier probabilities for samples $g(Z)$. For all metrics, we used Monte-Carlo estimates of expected values, based on 10,000 samples from $g(Z)$.
Expected nearest-neighbour distance ({\bf ENN}): To assess the similarity of generated samples with samples in the training set, we determined the
expected value of the Euclidean distance between samples from $g(Z)$ and their nearest neighbor in the train set. 
LeNet score ({\bf LS}): Similar to the Inception score proposed in \cite{Goodfellow_modified_GAN_2016},
we computed the exponential of the expected Kullback-Leibler divergence between the predicted class probabilities for samples $g(Z)$, and the frequency of classes (=digits) in the MNIST train set.
Jensen-Shannon divergence ({\bf JSD}): We estimated the JSD between the unknown data distributions and $g(Z)$ by computing (\ref{eq:Kernel_GANs_parameter}) over the MNIST test set and samples produced by $g(Z)$.
We also report the corresponding values {\bf JSD-F} and {\bf JSD-S} of the first and second term in (\ref{eq:Kernel_GANs_parameter}). Maximum Mean Discrepancy ({\bf MMD}): Finally, we also report the MMD statistic  (\cite{gretton2012kernel}).

Table \ref{tab:class_mean2} shows a comparison of different generators.
{\bf GMMN} and {\bf GMMN-AE} are the data- and code-space Generative Moment Matching Networks (GMMN) proposed in \cite{li2015generative}.
The numbers in the {\it MNIST Test} column are obtained by using the MNIST test set instead of generated samples; hence they can be regarded as the performance of an ideal generator, with optimal trade-off between fidelity (EE), diversity (ENN, LS),
and overall consistency (JSD). In this regard, we found DC performed the best among all trained generators: it achieved the lowest EE, comparable ENN, and the closest LS in comparison with {\it MNIST Test}. FC-FS and GMMN-AE also achieved high fidelity, but seemed to exhibit less diversity. Interestingly, the first and second term of the JSD were observed to be imbalanced for these models.
We hypothesize that keeping JSD-F and JSD-S balanced during training is key to obtaining generators with good generalization capacity.

\paragraph{CIFAR10.} We computed the Inception score (\cite{Goodfellow_modified_GAN_2016}) mean and standard deviation for 5$\times$ 10k samples obtained from a Kernel GAN that was trained for 5,000 iterations. The score for held-out CIFAR10 images (which can be regarded as gold standard) was 11.95 ($\pm$ .20). Kernel GANs yielded a score of 4.22 ($\pm$ .02), which is significantly higher than the scores for
GMMN-AE and GMMN (3.94 $\pm$ .04 and  3.47 $\pm$ .03, respectively), but lower than for MMD-GAN (6.17 $\pm$ .07, see \cite{li2017mmdgan}). The latter finding can be explained by the fact that we did not optimize hyperparameters and regularization weights for Kernel GANs, which could lead to further improvements in future work.

\section{CONCLUSIONS}\label{sec:Conclusions}

We established a rigorous framework for analyzing statistical properties of Generative Adversarial Network training.
To overcome potential pathologies (in particular, vanishing gradients),
we introduced a novel training objective, which can be regarded as minimizing a non-parametric estimate
of the Jensen-Shannon Divergence. We analyzed its asymptotic properties and showed its
practical applicability.

We see several directions for future work: 1) Advance the design of optimal kernels and strategies for annealing the bandwidths. 
2) Further analyze statistical properties of the proposed training objective, in particular, the effect of the regularizer.
3) Investigate the effect of imbalances between the first and second term in the training objective; we believe this could lead to the design of adaptive training protocols which ensure both fidelity and diversity of generator samples.

\bibliography{sample}

\newpage

\section*{A.1: Proofs}

\paragraph{Proof of Theorem \ref{theo:measure_preservation}:}
This is an immediate consequence of Theorem 1 in \citet{Schoenfeld_1975}, which builds on the classical result from Hahn-Mazurkiewicz that a metric space is the continuous image of the unit interval if and only if the space is compact, connected and locally connected.  \hfill $\square$

\paragraph{Proof of Theorem \ref{theo:population}:}
The arguments are analogous to the proofs of Proposition 1 and Theorem 1 in \citet{Goodfellow_GAN_original_paper_2014}, using the change-of-variable formula for general pushforward measures (\citet{Bogachev_2007}). \hfill $\square$

\paragraph{Proof of Theorem \ref{theo:optimal_discriminator}:}
This is an immediate consequence of Urysohn's Lemma (Section 15 in \citet{Willard_1970}), and the assumption that $\cX$ is a compact Hausdorff space. \hfill $\square$

\paragraph{Proof of Theorem \ref{theo:empirical_case}:}
We first note that 
\begin{eqnarray*}
V_n(d,g) &=& \int_{\cX} \Big[ \log(d(x)) p_n(x)  \,+\, \log(1-d(x)) p^{(g)}_n(x)  \Big] \mbox{\rm d}\sharp(x)
\end{eqnarray*}
where $\sharp$ denotes the counting measure on $(\cX,\cA)$, and $p_n$ and $ p^{(g)}_n$ the $\sharp$-densities 
of $X_n^\ast$ conditional on $X_1$, \ldots, $X_n$, and of  $g(Z_n^\ast)$ conditional on $Z_1$, \ldots, $Z_n$, respectively:
\begin{eqnarray}\label{proof:theo_empirical_case}
p_n(x) \,=\, \frac{1}{n}\sum_{i=1}^n \I(X_i=x) &\mbox{ and }&
 p^{(g)}_n(x) \,=\, \frac{1}{n}\sum_{i=1}^n \I(g(Z_i)=x)
\end{eqnarray}
for $x\in\cX$. Hence, completely analogous to Theorem \ref{theo:population}, we obtain
that any $d_n\in\cD$ maximizing $V_n(d,g)$ has the form
\begin{eqnarray*}
d_n(x) &=& \frac{p_n(x)}{p_n(x) + p^{(g)}_n(x)}
\end{eqnarray*}
for $x\in\{X_1,\ldots,X_n\}\cup\{g(Z_1),\ldots,g(Z_n) \}$, and hence the result in (\ref{eq:optimal_empirical_discriminator}) follows.
Again completely analogous to Theorem \ref{theo:population}, any generator $g_n\in\cG$ minimizing the objective {\rm (\ref{eq:empirical_objective})} for $d=d_n$ is such that $p_n(x) = p^{(g)}_n(x)$ for all $x\in\{X_1,\ldots,X_n\}\cup\{g(Z_1),\ldots,g(Z_n) \}$, which holds if and only if $\{g_n(Z_1),\ldots,g_n(Z_n) \}=\{X_1,\ldots,X_n\}$. \hfill $\square$

\paragraph{Proof of Theorem \ref{theo:optimal_modified_empirical_discriminator}:}
We note that 
$X_n^\ast+\epsilon$ conditional on $X_1$, \ldots, $X_n$ and $g(Z_n^\ast)+\epsilon$ conditional on $Z_1$, \ldots, $Z_n$
have the following $\mu$-densities:
\begin{eqnarray*}
(p_n\ast p^{(\epsilon)})(x) \,=\, \frac{1}{n}\sum_{i=1}^n p^{(\epsilon)}(x-X_i) &\mbox{ and }&
(p^{(g)}_n\ast p^{(\epsilon)})(x) \,=\, \frac{1}{n}\sum_{i=1}^n p^{(\epsilon)}(x-g(Z_i))
\end{eqnarray*}
for $\mu$-almost every $x\in\cX$, with $p_n$ and $p^{(g)}_n$ as given in (\ref{proof:theo_empirical_case}).
Using the same arguments as in the proof of Theorem \ref{theo:empirical_case}, we obtain both statements of this theorem. \hfill $\square$

\paragraph{Proof of Theorem \ref{theo:consistency}:}
Let $\theta$ and $\varphi$ be fixed.
First we note that the kernel $K$, the densities $p$, $p^{(\theta)}$ and the sequence $\sigma_n$ satisfy conditions (K2), (D2), (W2) in 
\citet{Gine_2002}. Consequently, by Theorem 3.3 in \cite{Gine_2002},
\begin{eqnarray*}
\lim_{n\to\infty} \sup_{x\in\cX} \big| \hat{p}_{n,\sigma_n}(x) - \E[ \hat{p}_{n,\sigma_n}(x)] \big| &=& 0
\end{eqnarray*}
$\P$-almost surely. Since $p$ is uniformly continuous and $K$ has compact support, it is easy to obtain from equation (1.3) in \cite{Gine_2002}:
\begin{eqnarray*}
\lim_{n\to\infty} \sup_{x\in\cX} \big| \E[ \hat{p}_{n,\sigma_n}(x)] - p(x) \big| &=& 0.
\end{eqnarray*}
Consequently, by the triangle inequality,
\begin{eqnarray}\label{eq:unif_converge_p}
\lim_{n\to\infty} \sup_{x\in\cX} \big| \hat{p}_{n,\sigma_n}(x) - p(x) \big| &=& 0
\end{eqnarray}
$\P$-almost surely. Using the same arguments, we obtain
\begin{eqnarray}\label{eq:unif_converge_p_theta}
\lim_{n\to\infty} \sup_{x\in\cX} \big| \hat{p}_{n,\sigma_n}^{(\theta)}(x) - p^{(\theta)}(x) \big| &=& 0
\end{eqnarray}
$\P$-almost surely. Now let $\kappa$ denote a finite upper bound both for $p$ and $p^{(\theta)}$ (which exists by our assumptions).
Note that $(r,s)\mapsto \log((r+\varphi)/(r+s+2\varphi))$ is uniformly continuous on $[0,\kappa]\times[0,\kappa]$. Hence, with (\ref{eq:unif_converge_p}) and (\ref{eq:unif_converge_p_theta}), we obtain
\begin{eqnarray*}
\lim_{n\to\infty} \sup_{x\in\cX} \Big|
 \log\frac{\hat{p}_{n,\sigma_n}(x)\,+\,\varphi}{\hat{p}_{n,\sigma_n}(x) \,+\, \hat{p}_{n,\sigma_n}^{(\theta)}(x)\,+\,2\varphi} \, - \,
 \log \frac{p(x)\,+\,\varphi}{p(x) \,+\,p^{(\theta)}(x)\,+\,2\varphi} \Big| &=& 0
\end{eqnarray*}
$\P$-almost surely. Consequently,
\begin{eqnarray}\label{theo:proof_consistency_eq1}
\left|\lim_{n\to\infty} \frac{1}{n} \sum_{i=1}^n \log \frac{\hat{p}_{n,\sigma_n}(X_i)\,+\,\varphi}{\hat{p}_{n,\sigma_n}(X_i) \,+\, \hat{p}_{n,\sigma_n}^{(\theta)}(X_i)\,+\,2\varphi}\,  - \, 
\lim_{n\to\infty} \frac{1}{n} \sum_{i=1}^n \log \frac{p(X_i)\,+\,\varphi}{p(X_i) \,+\,p^{(\theta)}(X_i)\,+\,2\varphi} \right| &=& 0
\end{eqnarray}
$\P$-almost surely. Moreover, by the Strong Law of Large Numbers,
\begin{eqnarray}\label{theo:proof_consistency_eq2}
\left| \lim_{n\to\infty} \frac{1}{n} \sum_{i=1}^n  \log \frac{p(X_i)\,+\,\varphi}{p(X_i) \,+\,p^{(\theta)}(X_i)\,+\,2\varphi} \, - \,
 \int_{\cX} p(x) \log  \frac{p(x)\,+\,\varphi}{p(x) \,+\, p^{(\theta)}(x)\,+\,2\varphi} \, \mbox{\rm d}\mu(x) \right| &=& 0
\end{eqnarray}
$\P$-almost surely. Hence, using (\ref{theo:proof_consistency_eq1}), (\ref{theo:proof_consistency_eq2}) and the triangle inequality,
we obtain
\begin{eqnarray*}
\left|\lim_{n\to\infty} \frac{1}{n} \sum_{i=1}^n \log \frac{\hat{p}_{n,\sigma_n}(X_i)\,+\,\varphi}{\hat{p}_{n,\sigma_n}(X_i) \,+\, \hat{p}_{n,\sigma_n}^{(\theta)}(X_i)\,+\,2\varphi}  \,  - \,
 \int_{\cX} p(x) \log  \frac{p(x)\,+\,\varphi}{p(x) \,+\, p^{(\theta)}(x)\,+\,2\varphi} \, \mbox{\rm d}\mu(x) \right| &=& 0
\end{eqnarray*}
$\P$-almost surely which is the desired result for the first term in (\ref{eq:theo_consistency}). The result for the second term follows by analogous arguments, which proves the theorem.

\section*{A.2: Asymptotically unbiased estimation of $\boldsymbol{p}$}

In order to establish $(ii)$, introduce the following function of $\theta$ and $\varphi$:
\begin{eqnarray*}
K(\theta,\varphi) &=& \int_{\cX} \Big[ (p(x)+\varphi) \log  \frac{p(x)\,+\,\varphi}{p(x) \,+\, p^{(\theta)}(x)\,+\,2 \varphi} \,\,+\,\, (p^{(\theta)}(x)+\varphi) \log \frac{p^{(\theta)}(x)\,+\,\varphi}{p(x)\,+\, p^{(\theta)}(x)\,+\,2 \varphi} \Big] \mbox{\rm d}\mu(x).
\end{eqnarray*}
Note that $K(\theta,\varphi)$ is the Jensen-Shannon Divergence (multiplied by $2\cdot(1+\varphi\cdot\mu(\cX))$) between the densities
\begin{eqnarray*}
\tilde{p}(x) \,:=\, \frac{p(x)+\varphi}{1+\varphi\cdot\mu(\cX)} &\mbox{ and }&
\tilde{p}^{(\theta)}(x) \,:=\, \frac{p^{(\theta)}(x)+\varphi}{1+\varphi\cdot\mu(\cX)}.
\end{eqnarray*}
Hence, $K(\theta,\varphi)$ is minimized for $\theta$ such that $\tilde{p}^{(\theta)}=\tilde{p}$ $\mu$-almost everywhere, which is
equivalent to $p^{(\theta)}=p$ $\mu$-almost everywhere.
Next, observe that $K(\theta,\varphi)$ is equal to the right-hand side in (\ref{eq:theo_consistency}), plus the following two terms:
\begin{eqnarray*}
K^{(1)}(\theta,\varphi) &:=&
\int_{\cX} \varphi \log  \frac{p(x)\,+\,\varphi}{p(x) \,+\, p^{(\theta)}(x)\,+\,2 \varphi}\, \mbox{\rm d}\mu(x), \\
K^{(2)}(\theta,\varphi) &:=&
\int_{\cX} \varphi \log \frac{p^{(\theta)}(x)\,+\,\varphi}{p(x)\,+\, p^{(\theta)}(x)\,+\,2 \varphi}\, \mbox{\rm d}\mu(x).
\end{eqnarray*}
Now suppose $\tilde{X}_1$, $\tilde{X}_2$, \ldots are sampled independently according to the probability density $x\mapsto\mu(x)/\mu(\cX)$, and consider the estimators
\begin{eqnarray*}
K^{(1)}_n(\theta,\sigma,\varphi) &:=&
\frac{\mu(\cX)}{n} \sum_{i=1}^n\log \frac{\hat{p}_{n,\sigma}(\tilde{X}_i)\,+\, \varphi}{\hat{p}_{n,\sigma}(\tilde{X}_i) \,+\, \hat{p}_{n,\sigma}^{(\theta)}(\tilde{X}_i)\,+\,2 \varphi}, \\
K^{(2)}_n(\theta,\sigma,\varphi) &:=&
\frac{\mu(\cX)}{n} \sum_{i=1}^n\log \frac{\hat{p}_{n,\sigma}^{(\theta)}(\tilde{X}_i)\,+\, \varphi}{\hat{p}_{n,\sigma}^{(\theta)}(\tilde{X}_i) \,+\, 
\hat{p}_{n,\sigma}(\tilde{X}_i)\,+\,2 \varphi}.
\end{eqnarray*}
Under the same assumptions as in Theorem \ref{theo:consistency}, we obtain
\begin{eqnarray*}
\lim_{n\to\infty} \big( K^{(1)}_n(\theta,\sigma_n,\varphi) + K^{(2)}_n(\theta,\sigma_n,\varphi) \big)
&=& K^{(1)}(\theta,\varphi) + K^{(2)}(\theta,\varphi)
\end{eqnarray*}
$\P$-almost surely. Hence, $K_n(\theta,\sigma_n,\varphi)  + K^{(1)}_n(\theta,\sigma_n,\varphi)  + K^{(2)}_n(\theta,\sigma_n,\varphi) $ converges to a limit which is minimized by $\theta$ such that the generator $g^{(\theta)}(Z)$ recovers $X$.

\section*{A.3: Implementation details}

\textbf{MNIST Autoencoder.} The autoencoder for Feasture Space (FS) based Kernel GANs of MNIST were trained to yield 32-dimensional feature vectors of images, as suggested in \cite{li2015generative}. The architecture used was: 784-1024-sigmoid-32-sigmoid-32-sigmoid-1024-sigmoid-784. The model was trained with cross-entropy as the reconstruction loss, dropouts in the encoder layers, and Adam \cite{kingma2014adam} for optimization.  

\textbf{MNIST Generators.} The achitectures used for the generators were:
\begin{itemize}
    \item \textbf{FC}: 10-64-relu-256-relu-256-relu-1024-relu-784-sigmoid (\cite{li2015generative}), \\
    Kernel: mixture of RBF ($\sigma\in\{100.0, 50.0, 10.0, 5.0, 1.0, 0.5, 0.1, 0.05, 0.01\}$).
    \item \textbf{FC-FS}: 10-64-relu-256-relu-256-relu-32-sigmoid (\cite{li2015generative}) \\
    Kernel: mixture of RBF ($\sigma\in\{1.0, 0.5, 0.1, 0.05, 0.01\}$).
    \item \textbf{DC}: the DCGAN (\cite{radford2015unsupervised}) architecture with dimension as 64. \\
    Kernel: mixture of RBF ($\sigma\in\{100.0, 50.0, 10.0, 5.0, 1.0, 0.5, 0.1, 0.05, 0.01\}$).
\end{itemize}

Training was performed with a minibatch size of $n=1,000$ (cf Algorithm \ref{algo:KGAN}), and RMSProp \cite{tieleman2012lecture} (learning rate of 0.001) for optimization. 

\textbf{Classifier for LeNet score.} A LeNet-like classifier was trained with the following architecture: (28,28,1)-conv(32,(3,3))-relu-maxpool(2,2)-conv(64,(3,3))-relu-maxpool(2,2)-fc(128)-relu-10-softmax. Training was performed using Adam and with dropout for regularization. 

\textbf{Generative Moment Matching Networks.} We trained the data-space and code-space networks of \cite{li2015generative}, which they define as GMMN and GMMN-AE respectively. We used a mixture of RBF kernels (GMMN: $\sigma\in\{100.0, 50.0, 10.0, 5.0, 1.0, 0.5, 0.1, 0.05, 0.01\}$, GMMN-AE: $\sigma\in\{1.0, 0.5, 0.1, 0.05, 0.01\}$).

\textbf{CIFAR10 and CelebA.} We trained models based on DCGAN (\cite{radford2015unsupervised}) architecture for both CIFAR10 and CelebA. Similar to \cite{li2017mmdgan}, we trained a network for $f$ which was modelled as an encoder of a convolutional autoencoder. The encoding dimension was fixed to 100. A mixture of RBF kernels ($\sigma\in\{1.0, 2.0, 4.0, 8.0, 16.0\}$) was used as kernels in equation \ref{eq:Kernel_GANs_parameter}. As mentioned in section \ref{sub:scaling_kernel_gans}, training of $f$ was regularised with autoencoder reconstruction loss (with a weight of 100). Additionally, the weights of the network $f$ were clipped to a range of $\{-0.01,0.01\}$ after every update. This setup was identitical to the one used in \cite{li2017mmdgan}\footnote{https://github.com/OctoberChang/MMD-GAN}. For computing Inception Score, we used the script provided in the work of \cite{Gulrajani_et_al_2017} \footnote{https://github.com/igul222/improved\_wgan\_training/blob/master/tflib/inception\_score.py}

Hyperparameters including the model architecture, values of kernel bandwidth were not optimised for any of the experiments. Further, we also believe that longer schedules for optimisation can affect the performance of generators. 

\section*{A.4: Experiments }

\begin{figure}[H]
 \centering  
    \begin{minipage}{0.3\textwidth}
        \centering
        \includegraphics[scale=0.3]{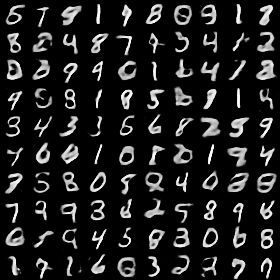}
        \caption{DC}
        \label{fig:mnisttest_DC}
    \end{minipage}
        \begin{minipage}{0.3\textwidth}
        \centering
        \includegraphics[scale=0.3]{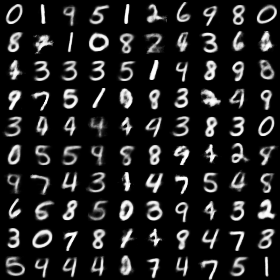}
        \caption{FC-FS}
        \label{fig:mnisttest_FC_FS}
    \end{minipage}
    \begin{minipage}{0.3\textwidth}
        \centering
        \includegraphics[scale=0.3]{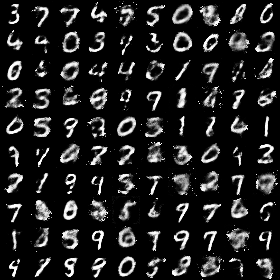}
        \caption{FC}       
        \label{fig:mnisttest_FC}
    \end{minipage}
\end{figure}

\begin{figure}[H]
    \centering
    \begin{minipage}{0.3\textwidth}
        \centering
        \includegraphics[scale=0.3]{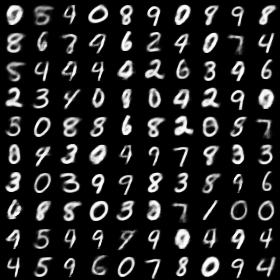}
        \caption{FC-FS for RBF kernel with bandwidth $\sigma$=0.01}
        \label{fig:mnisttest_FC_FS_RBF}
    \end{minipage}
    \begin{minipage}{0.3\textwidth}
        \centering
\includegraphics[scale=0.3]{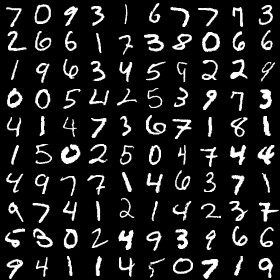}
        \caption{MNIST test set}
        \label{fig:mnisttest_MNIST}
    \end{minipage}
\end{figure}

\textbf{MNIST generated samples.}
Figure \ref{fig:mnisttest_DC}--\ref{fig:mnisttest_FC_FS_RBF} show the generator samples for MNIST. For comparison, we include a sample from the MNIST test set in Figure \ref{fig:mnisttest_MNIST}.
 

\textbf{Effect of the kernel bandwidth.}
We conducted experiments with the FC-FS architecture and simple RBF kernels with different bandwidths. Figure \ref{fig:mnisttest_FC_FS_RBF} shows
samples generated for the MNIST dataset. As can be seen, the sample quality is comparable to FC-FS, however, some digits appear to be over- and under-represented, respectively. 
Table \ref{tab:class_mean1} shows the EE and EEN metrics for on the MNIST dataset for different bandwidths. We note that smaller bandwidths result in generator samples with lower EE. 
Below we show EE values for samples generated over 2-dimensional manifolds in the latent space, illustrating
that lower EE stems both from generated samples with higher visual fidelity, and sharper transitions between low-entropy regions in the latent space.
Similarly, also the EEN values decrease with the bandwidth. This can be regarded as a loss of diversity, as generated samples become more and more similar to instances in the training set.

\begin{table}
  \caption{Effect of the kernel bandwidth $\sigma$ on the FC-FS model}
  \label{tab:class_mean1}
  \centering
\begin{tabular}{lrrrrrrr}
\toprule
                        &   $\sigma=5.0$     &   $\sigma=1.0$     &   $\sigma=0.5$     &   $\sigma=0.1$     &  $\sigma=0.05$  &  $\sigma=0.01$   & sum-RBF  \\
\midrule
\textbf{EE}    & 2.117     & 1.388     &  1.096    &   1.079   &  0.783    &  0.406    & 0.365     \\
\textbf{ENN}    & 28.973 & 21.075   &  18.268  &   17.295 &  16.272  &  15.378  & 14.730   \\
\bottomrule
\end{tabular}
\end{table}

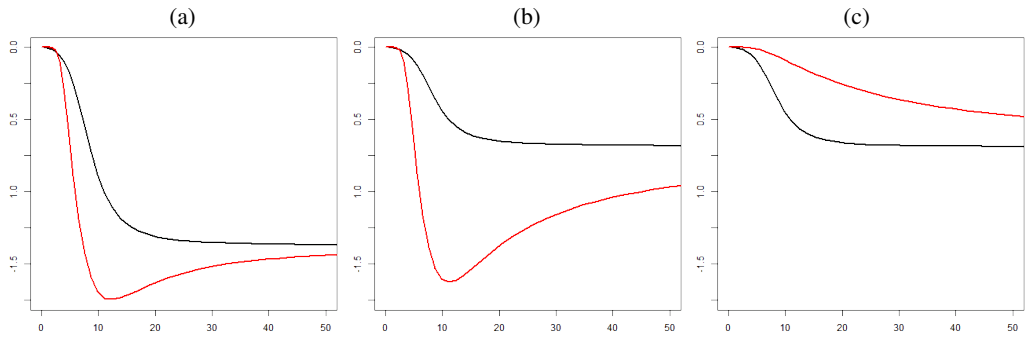
\begin{figure*}
\begin{picture}(400,150)
\put(65,130){\small (a)}
\put(195,130){\small (b)}
\put(320,130){\small (c)}
\put(-5,0){\includegraphics[width=5cm]{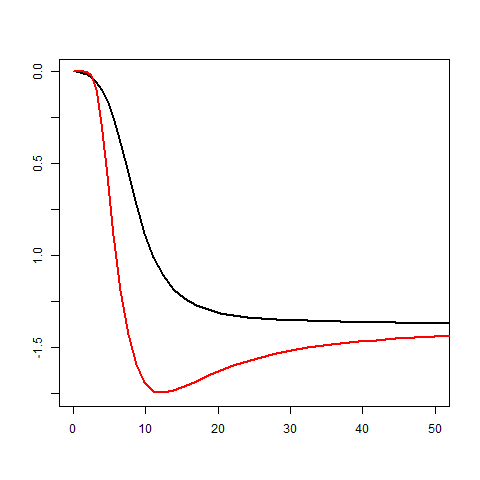}}
\put(125,0){\includegraphics[width=5cm]{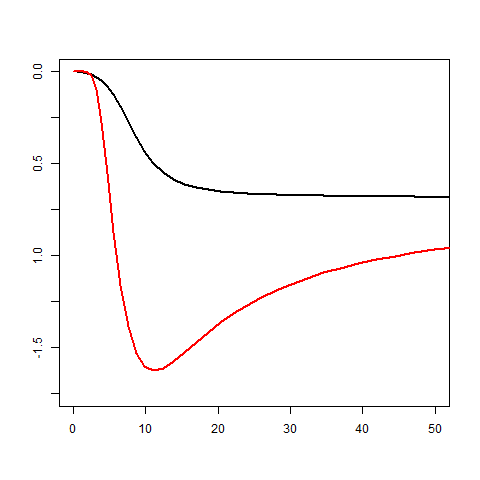}}
\put(255,0){\includegraphics[width=5cm]{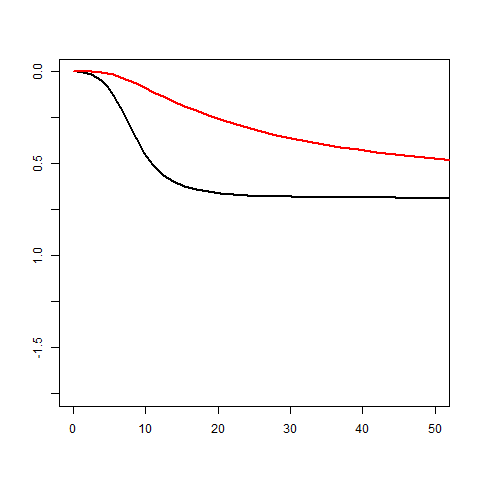}}
\end{picture}
\caption{Effect of the kernel bandwidth $\sigma$ on the values of the training objective function (\ref{eq:Kernel_GANs_parameter}). 
The black lines show the objective function values for an ``ideal'' generator where $\{X_1$, \ldots, $X_n\}$ and $\{g(Z_1)$, \ldots, $g(Z_n)\}$ are two disjoint subsets of the MNIST training set (for $n=100$). The red line shows the objective function values for a generator which produces $g(Z_i)$ equal to the mean of $\{X_1$, \ldots, $X_n\}$. All objective function values are averaged over 100 random selections of MNIST subsets. The regularization paramter $\varphi$ is chosen zero everywhere.
(a) Values of the objective function (\ref{eq:Kernel_GANs_parameter}) depending on $\sigma$. (b) Values of the first term in (\ref{eq:Kernel_GANs_parameter}).
(b) Values of the second term in (\ref{eq:Kernel_GANs_parameter}). }
\label{fig: bandwidth analysis}
\end{figure*}

We observed that, for very large values of the kernel bandwidth $\sigma$, the generated samples have a tendency to
collapse to the mean of the training instances. Figure \ref{fig: bandwidth analysis} analyzes the behaviour of the training objective in (\ref{eq:Kernel_GANs_parameter})
as a function of $\sigma$. We compare two different generators: an ``ideal'' one (black line), which is able -- given 100 samples from the MNIST training set, to produce 100 different samples. The red line shows objective function values of a generator which simply produces the mean of the 100 training samples. As can be seen in (a), for large values of $\sigma$ the latter generator performs better with regard to the training objective. For small $\sigma$ values, the order is reverse. Interestingly, for large $\sigma$ the objective function values of the generator producing the mean is below $-\log(4)$. As (b) and (c) show, this is due to a large imbalance of the first and second term in the training objective. Hence, we hypothesize that the training of meaningful generators should not only aim to minimize (\ref{eq:Kernel_GANs_parameter}), but also aim to keep the first and second term in (\ref{eq:Kernel_GANs_parameter}) balanced. A deeper investigation of the trade-off between these two terms will be future work.

\begin{figure}[H]
    \centering
        \includegraphics[scale=0.375]{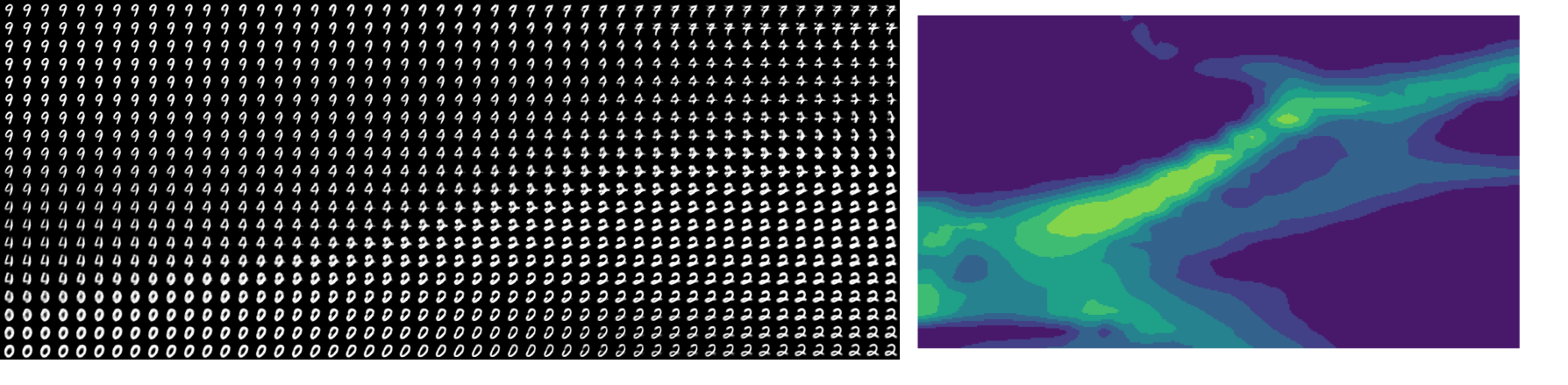}
        \caption{Generated images on 2-dimensional manifold and corresponding entropy carpet (FC-FS)}
        \label{fig:entropy_fcfs}
\end{figure}

\begin{figure}[H]
    \centering
        \includegraphics[scale=0.375]{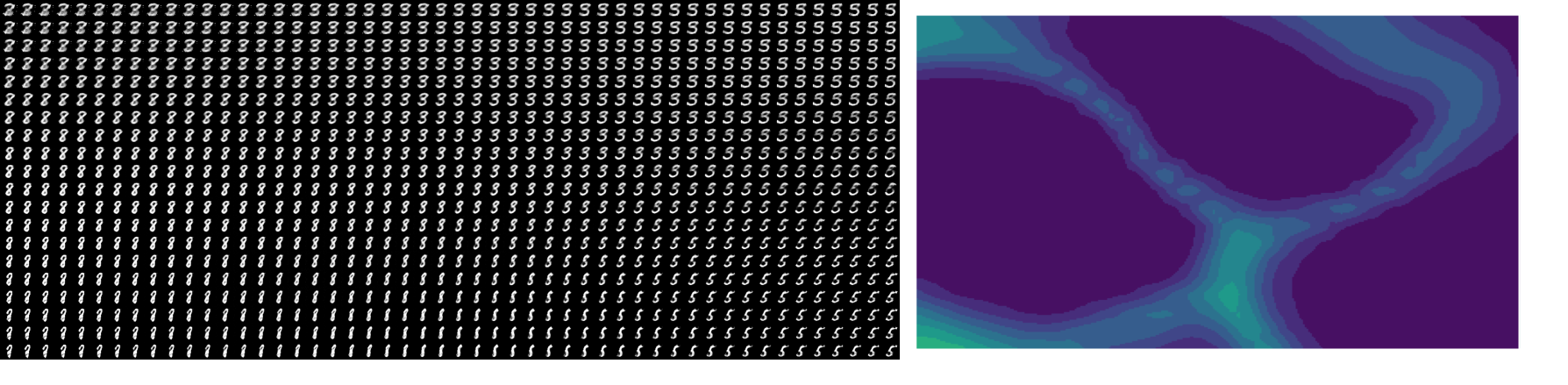}
        \caption{Generated images on 2-dimensional manifold and corresponding entropy carpet (FC)}
        \label{fig:entropy_fc}
\end{figure}

\begin{figure}[H]
    \centering
        \includegraphics[scale=0.375]{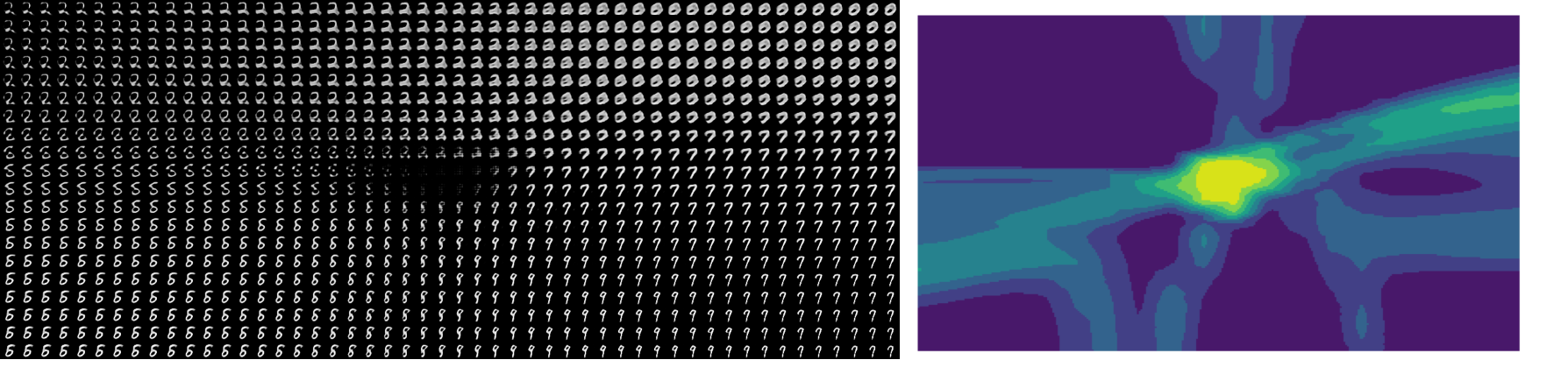}
        \caption{Generated images on 2-dimensional manifold and corresponding entropy carpet (DC)}
        \label{fig:entropy_conv}
\end{figure}

\textbf{Entropy carpets.} Given a trained GAN $g$ and points $z_1$, $z_2$ in the latent space $\cZ$, it is often instructive to inspect the generated images $g((1-x)\cdot z_1+x \cdot z_2)$ for $x\in[0,1]$.
Recently, \citet{Dinh_Real_NVP_2016} proposed an angle-based 2-dimensional manifold interpolation between four points $z_1,z_2,z_3,z_4\in\cZ$.
Here we explore more conventional convex combinations, given by
\begin{eqnarray*}
\{ x\cdot y\cdot z_1 + (1-x)\cdot y\cdot z_2 + x\cdot(1-y)\cdot z_3 + (1-x)\cdot(1-y)\cdot z_4: \ \ x,y\in[0,1]\}.
\end{eqnarray*}
In practice, we let $x$ and $y$ vary along a mesh grid of a unit-length square. Figure \ref{fig:entropy_fcfs}-\ref{fig:entropy_conv} display the image manifolds generated by three different models, along with the ``entropy carpets'' which show the entropy of a probabilistic classifier (we use LeNet, see above) at each point of
the manifold. Bright colors correspond to high-entropy regions, dark colors to low entropy. Entropy carpets can be regarded as a semi-qualitative-semi-quantitative way to relate the manifolds of generated images to the Expected Entropy (EE) metric reported in Tables \ref{tab:class_mean1} and \ref{tab:class_mean2}.
Of particular interest are the inter-digit transitions which coincide with high entropy. Typically, the output of the generator is uninterpretable in those regions. Hence, an ideal generator should have as few and as sharp transitions between different modes as possible.

\begin{figure}[H]
    \centering
    \begin{minipage}{0.3\textwidth}
        \centering
        \includegraphics[scale=0.3]{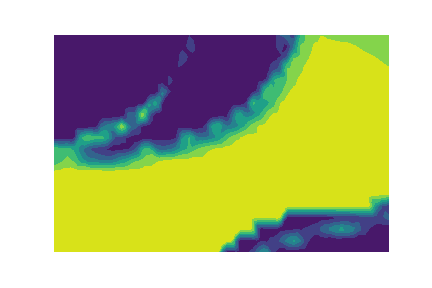}
        \caption{Entropy carpet for FC-FS-RBF(0.5)}
        \label{fig:mnisttest-ec-1}
    \end{minipage}
    \begin{minipage}{0.3\textwidth}
        \centering
        \includegraphics[scale=0.3]{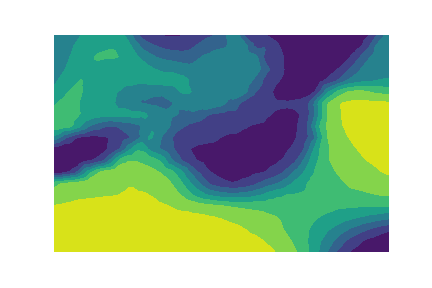}
        \caption{Entropy carpet for FC-FS-RBF(0.05)}
        \label{fig:mnisttest-ec-2}
    \end{minipage}
    \begin{minipage}{0.3\textwidth}
        \centering
        \includegraphics[scale=0.3]{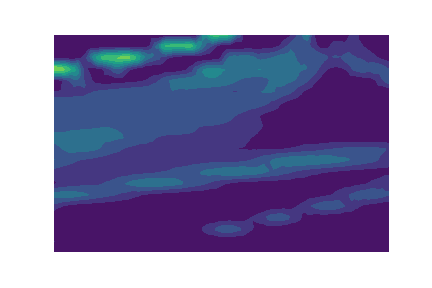}
        \caption{Entropy carpet for FC-FS-RBF(0.01)}
        \label{fig:mnisttest-ec-3}
    \end{minipage}
\end{figure}

Figures \ref{fig:mnisttest-ec-1}-\ref{fig:mnisttest-ec-3} show the entropy carpets for the FC-FS model trained with bandwidths $\sigma=0.5, 0.05, 0.01$.
Beyond the Expected Entropy metrics reported in Tables \ref{tab:class_mean1}, the entropy carpets give an idea of the fraction of points $\cZ$ that result in meaningful versus non-meaningful images. Interestingly, even for $\sigma=0.5$, significant parts of $\cZ$ result in high-fidelity images; however, there are large areas in between the modes in which the generator only generates ``noise'' from the point-of-view of the classifier. .

\end{document}